\documentclass[acmtog]{acmart}

\usepackage{booktabs} 
\usepackage{graphics}

\usepackage[ruled]{algorithm2e} 

\SetAlFnt{\small}
\SetAlCapFnt{\small}
\SetAlCapNameFnt{\small}
\SetAlCapHSkip{0pt}

\acmYear{2024}
\acmMonth{12}

\copyrightyear{2024}
\acmYear{2024}
\setcopyright{acmlicensed}\acmConference[SA Conference Papers '24]{SIGGRAPH Asia 2024 Conference Papers}{December 3--6, 2024}{Tokyo, Japan}
\acmBooktitle{SIGGRAPH Asia 2024 Conference Papers (SA Conference Papers '24), December 3--6, 2024, Tokyo, Japan}
\acmDOI{10.1145/3680528.3687609}
\acmISBN{979-8-4007-1131-2/24/12}

\begin{document}
\newcommand{\TXJ}[1]{{\color{blue}#1}}
\title{Decoupling Contact for Fine-Grained Motion Style Transfer}
\author{Xiangjun Tang}
\orcid{0000-0001-7441-0086}
\affiliation{%
 \institution{State Key Lab of CAD\&CG, Zhejiang University
 }
 \city{Hangzhou}
  \country{China}
 }
 \email{xiangjun.tang@outlook.com}

\author{Linjun Wu}
\orcid{0000-0002-1988-0090}
\affiliation{%
 \institution{State Key Lab of CAD\&CG, Zhejiang University
 }
  \city{Hangzhou}
  \country{China}
 }
 \email{12321232@zju.edu.cn}

\author{He Wang}
\orcid{0000-0002-2281-5679}
\affiliation{%
 \institution{Department of Computer Science and UCL Centre for Artificial Intelligence, University College London}
  \city{London}
  \country{United Kingdom}
 }
 \email{he_wang@ucl.ac.uk}

\author{Yiqian Wu}
\orcid{0000-0002-2432-809X}
\affiliation{%
 \institution{State Key Lab of CAD\&CG, Zhejiang University}
  \city{Hangzhou}
  \country{China}
 }
 \email{onethousand1250@gmail.com}

\author{Bo Hu}
\orcid{0000-0002-6599-7249}
\affiliation{%
 \institution{Tencent Technology Co., Ltd.}
 \city{Shenzhen}
  \country{China}
}
 \email{corehu@tencent.com}

\author{Songnan Li}
\orcid{0000-0001-6931-4129}
\affiliation{%
 \institution{Tencent Technology Co., Ltd.}
  \city{Shenzhen}
  \country{China}
}
 \email{sunnysnli@tencent.com}

\author{Xu Gong}
\orcid{0000-0003-3900-2903}
\affiliation{%
 \institution{Tencent Technology Co., Ltd.}
  \city{Shenzhen}
  \country{China}
}
 \email{xugong@tencent.com}

\author{Yuchen Liao}
\orcid{0000-0002-9008-3609}
\affiliation{%
 \institution{Tencent Technology Co., Ltd.}
  \city{Shenzhen}
  \country{China}
}
 \email{bluecatliao@tencent.com}

\author{Qilong Kou}
\orcid{0000-0002-5222-7069}
\affiliation{%
 \institution{Tencent Technology Co., Ltd.}
  \city{Shenzhen}
  \country{China}
}
 \email{rambokou@tencent.com}

\author{Xiaogang Jin}
\authornote{Corresponding author}
\orcid{0000-0001-7339-2920}
\affiliation{%
 \institution{State Key Lab of CAD\&CG, Zhejiang University; ZJU-Tencent Game and Intelligent Graphics Innovation Technology Joint Lab
 }
 \city{Hangzhou}
  \country{China}
}
 \email{jin@cad.zju.edu.cn}

\renewcommand\shortauthors{Tang et al.}

\begin{abstract}
Motion style transfer changes the style of a motion while retaining its content and is useful in computer animations and games.
Contact is an essential component of motion style transfer that should be controlled explicitly in order to express the style vividly while enhancing motion naturalness and quality. However, it is unknown how to decouple and control contact to achieve fine-grained control in motion style transfer.

In this paper, we present a novel style transfer method for fine-grained control over contacts while achieving both motion naturalness and spatial-temporal variations of style. Based on our empirical evidence, we propose controlling contact indirectly through the hip velocity, which can be further decomposed into the trajectory and contact timing, respectively. To this end, we propose a new model that explicitly models the correlations between motions and trajectory/contact timing/style, allowing us to decouple and control each separately. Our approach is built around a motion manifold, where hip controls can be easily integrated into a Transformer-based decoder. It is versatile in that it can generate motions directly as well as be used as post-processing for existing methods to improve quality and contact controllability. In addition, we propose a new metric that measures a correlation pattern of motions based on our empirical evidence, aligning well with human perception in terms of motion naturalness. 
Based on extensive evaluation, our method outperforms existing methods in terms of style expressivity and motion quality.

\end{abstract}

%
%


%
%


\begin{teaserfigure}
\centering
  \includegraphics[width=\textwidth]{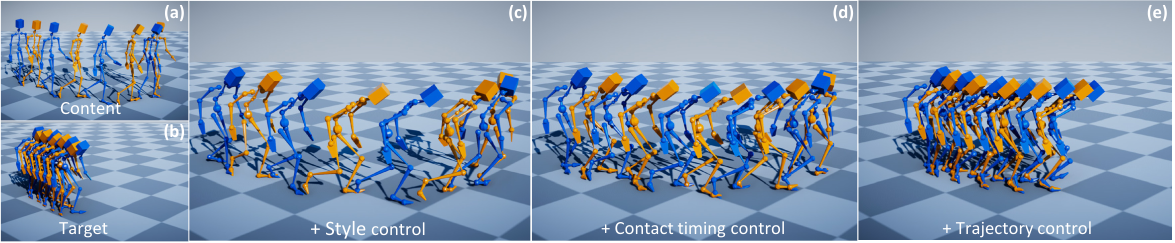}
  \caption{
  {
  Our method can independently control style, contact timing, and trajectory, allowing for fine-grained motion style transfer. Given a content motion (a) and an "old man" style (bending, fast pace, and slow speed) target motion (b), our approach allows for the gradual addition of "style" (c), "contact timing" (d), and "trajectory" (e) of the target motion to the content, which previous methods could not achieve. The result in (c) depicts the target motion's bending pose; the result in (d) depicts more frequent contact within the same time duration, indicating a faster pace of the character; and the result in (e) depicts a slower speed, precisely replicating the entire "old man" style. We show every frame in which either foot makes contact with the ground. 
  }
  }
  \Description{}
  \label{fig:teaser_overview}
\end{teaserfigure}

\maketitle
\newcommand{\HW}[1]{{\textcolor{red}{(#1)}}}

\definecolor{autumnorange}{rgb}{0.87, 0.61, 0.33}
\newcommand{\onethous}[1]{{\color{autumnorange}#1}}

\section{Introduction}
Motion style transfer has important applications in computer animation and gaming by reducing laborious and costly motion capture and empowering artists in creation. Typically, motion style transfer is accomplished by separating the style from the content so that the style can be transferred to a different content~\cite{aberman2020unpaired,Jiang_motion_puzzle,holden_deep_2016,park2021diverse,yumer2016spectral}. 
However, in character motions, it is difficult to distinguish between content and style, which can sometimes lead to ambiguity~\cite{song2023finestyle}.

Given such an ambiguity, existing work interprets content and styles differently~\cite{yumer2016spectral,Jiang_motion_puzzle,song2023finestyle,amaya1996emotion}, which can be broadly divided into supervised styles and unsupervised styles.  Supervised style uses human-labeled actions and treats style as spatial-temporal variations~\cite{ribet2019survey} (amplitude, speed, and pose, for example) of an action. Walking motions, for example, may include various stepping strategies (stride, stroll, tramp, march) with varying strides, stepping frequency, knee height during stepping, and so on, all of which are treated as style. 
Unsupervised style relies on motion similarity rather than human labeling. Clustered motions based on similarity (for example, in the data space or some latent space) can be viewed as variations on the same content \cite{Jiang_motion_puzzle}. Despite the cutting-edge performance of deep learning under both strategies \cite{aberman2020unpaired,Jiang_motion_puzzle,song2023finestyle}, it is safe to say that a complete separation of motion style and content is difficult to achieve. 

If a complete decoupling is difficult, we ask a question: what information cannot be decoupled completely and is it important in motion style transfer? We find contact is such an element and it is crucial. In locomotion, contact is multi-faceted including timing, duration, frequency, pattern, location, etc. It is tightly coupled with both content and style. Although most existing work treats contact as part of the content~\cite{aberman2020unpaired,Jiang_motion_puzzle,yumer2016spectral,unuma1995fourier},  it has recently been recognized as being closely related to style as well~\cite{aberman2020unpaired,dong2020adult2child}. However, it is unknown how to control contact to achieve fine-grained control in motion style transfer. As will be demonstrated later, not explicitly controlling contact frequently results in limited expressiveness of certain styles or even compromising the content itself, e.g. unnatural motions.

To this end, we present a novel style transfer method based on fine-grained contact control that achieves both motion naturalness and spatial-temporal variations of style. Given a source motion containing the action's main content and another motion containing the desired style, our approach allows for the interpolation of the multi-faceted contact information between them, resulting in controllable and smooth transitions. However, because contact is so closely linked to both content and style, naive direct contact control results in unnatural and uncontrollable motions. We propose to control contact indirectly via the hip velocity, which can be further decomposed into high-level and low-level features based on empirical statistical evidence from the data. We discovered that high-level features can largely determine the root trajectory and thus the spatial aspect of contact, such as locations, whereas low-level features can govern the timing. They work together to provide fine-grained control during style transfer.

To achieve the above, we propose a new model to learn a motion manifold where the correlations between motions and other factors (trajectory, contact timing and style) are explicitly modeled. Specifically, we first employ different neural networks to encode the hip trajectory, contact timing of the content motion, and style features of the style motion separately. We then fuse them to control the trajectory and contact timing, as well as a latent variable that captures the style features. Following that, we obtain a controllable motion manifold into which hip controls can be easily added for final motion synthesis using a Transformer-based decoder. 
The core of our approach is the motion manifold, which, as demonstrated later, can enhance motion quality and contact controllability. It is versatile in that it can directly generate motions, and can also be used as post-processing for existing methods \cite{aberman2020unpaired,Jiang_motion_puzzle}. In addition, we propose \textit{Contact Precision-Recall}, which measures the match between the synthesized contact and hip velocity based on our empirical observation of their high correlation. This metric aligns better with human perception in measuring motion quality than Fr\'{e}chet Motion Distance (FMD) and foot skating metrics.
We demonstrate that our method outperforms existing methods in terms of style expressivity and motion quality through extensive validation. 

The contributions of our work can be summarized as follows:
\begin{itemize}
    \item  A novel method for fine-grained control of motion style transfer, which improves the expressiveness and naturalness of motion style transfer.
    \item A new transformer-based model for motion manifold based on empirical evidence that allows us to control contacts by hip velocity, resulting in more natural and refined motions.
    \item Our manifold can be combined with existing motion transfer methods to improve motion quality and controllability. 
\end{itemize}

\section{Related work}
\subsection{Controllable Motion Generation}
Our method allows for contact control via hip velocity, creating a continuous space for contact editing. A related field is controllable motion generation, which can be formulated as motion planning problems \cite{wang_harmonic_2013,Safonova07constructionand,beaudoin2008motion,levine2012continuous, wang_energy_2015}. However, motion planning problems require complex optimization~\cite{chai_constraint-based_2007} and frequently lead to slow computation.
By searching in structured data, data-driven methods~\cite{Kovar_motion_2002,Min_motion_2012,shen_posture_2017,arikan_interactive_2002,kim2023interactive} can avoid slow optimizations but require unaffordable memory space to cover diverse control situations.
Deep neural networks can leverage compressed data representations~\cite{holden_learned_2020}.
One method for incorporating controllability is to include constraints as regularization in the loss function~\cite{chiu2019action,martinez2017human,Wang_STRNN_2019}. 
When different contacts are required, however, simply adding constraints will not yield high-quality results. Learning the conditional probability via a generative model, such as VAE \cite{petrovich2022temos,petrovich2021action,Chen_dynamic_2020,tang2022cvae,ling_character_2020,zhang2023learning}, GAN \cite{ahn2018text2action}, flows \cite{alexanderson2020style}, and diffusion models \cite{tevet2022diffusion,chen2023executing,alexanderson2023listen,Ao2023GestureDiffuCLIP,ghorbani2022zeroeggs}, can produce controlled natural motions. 
However, controlling contact or interpolating contact between two motions without affecting the style hasn't been thoroughly studied.


\subsection{Motion Style Transfer}
To achieve motion style transfer, early work aligns two motions to characterize their differences~\cite{hsu2005style}, or by modeling the style in frequency domains~\cite{unuma1995fourier,pullen2002motion,bruderlin1995motion,yumer2016spectral}, but they largely handle relatively small amounts of data. 
Recent methods relying on a large labeled dataset learn the mapping between two different domains~\cite{dong2020adult2child,almahairi2018augmented} or model the style directly as the common features across all motions with the same style label~\cite{xia2015realtime,Mason2022Style}. The style can be represented as one-hot embedding~\cite{smith2019efficient,park2021diverse,chang2022unifying} or style variable~\cite{brand2000style}. However, these representations lack the details of the given style sequence. 
Another strategy models the style as the variance of the latent vectors, using Gram Matrix~\cite{holden2017fast} or AdaIN~\cite{park2021diverse,aberman2020unpaired}. To extract the fine-grained style variance, the following methods~\cite{Jiang_motion_puzzle,jang2023mocha,song2023finestyle,kim2024most} incorporate the body-part level attention mechanism.
However, these methods do not handle temporal difference~\cite{dong2020adult2child} and hardly handle hip velocity correctly  
since hip velocity is coupled with both style and content.
Stylization is a related area of style transfer in which specific style motions are generated without involving the original motions. Autoregressively stylized motion generators \cite{Mason2022Style,Mason_2018_fewshot,tang2023rsmt,tao2022style,xia2015realtime} generate high-quality diverse motion while adhering to predefined constraints such as trajectory or keyframes. These methods, however, cannot incorporate the content of another motion.




\section{Methodology}

\subsection{Hip Velocity-Contact Timing Relationship}
\label{sec:observation}

\begin{figure}[h]
    \includegraphics[width=1.0\linewidth]{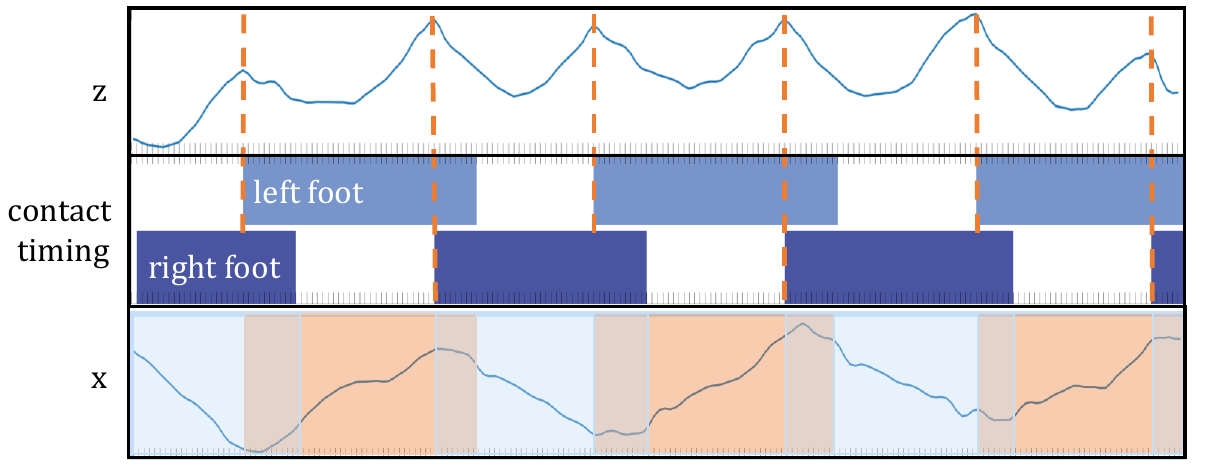}
    \caption{The diagram illustrates the correlation between the contact pattern and the hip speed of a walking sequence from the STYLE100 Dataset. The ``z'' axis of our frame points to the character's forward-facing direction and the ``x'' axis points to the character's left, both in the local coordinate system at the current frame. In the middle (contact timing), there are two rows of bars (light blue and blue). The top bars represent the left foot contact duration and the bottom ones represent the right foot.   The first row shows the ``z'' component of the hip velocity, in which there is a peak value when the foot makes contact with the ground. The curve in the third row depicts the ``x'' component of the hip velocity, with orange and light blue rectangles representing left and right foot contact duration, respectively. The ``x'' component of the hip velocity decreases (increase in negative ``x'' axis) during the right leg contact, and increases during the left leg contact.
    }
    \label{fig:observations}
\end{figure}

Explicitly controlling contact would enable fine-grained style transfer, which enhances both the style expressiveness and the content quality. A naive approach would be to constrain the contact velocity and position. However, since the contact is tightly coupled with both style and content, this would result in unnatural and uncontrollable motions, as demonstrated in the Appendix. To this end, we aim to find a proxy, which can be leveraged to control the contact and can also be easily decoupled from the style and the content. {Notably, while phase~\cite{Starke_deeppahse_2022} is a popular implicit motion representation related to contact, it is not an ideal choice for two reasons. First, it is relevant to both style and content~\cite{tang2023rsmt} and cannot be easily decoupled. Second, editing an implicit representation for flexible motion control is not straightforward. Instead,}
we find that \textit{hip velocity} might be a good proxy because it is loosely related to the characteristics that are essential for expressing the style or content, such as body movement and poses, but also easily controllable and correlated to the contact. As illustrated in Fig.~\ref{fig:observations}, although the hip speed magnitude is not explicitly related to the contact, the timing change of the hip speed trend such as the increasing trend to decreasing trend corresponds to the switch timing in contact.

To achieve the goal of controlling the contact by hip velocity, the remaining questions are whether the hip-contact relationship exists for other diverse motions and whether the hip velocity sequence of a motion is adequate for inferring the contact timing pattern. 
We conduct an experiment to explore these questions. Specifically,
we model the potential hip-contact relationship with a convolution neural network, denoted by $f_\delta(\cdot)$, trained on two different motion datasets.  
The network takes the hip velocity  $\mathbf{h} \in \mathcal{R}^{T\times 3}$ as input and predicts the contact states $c_t \in \mathcal{R}^{T\times 2}$ for two legs.
The experimental results on two datasets 
both demonstrate high prediction precision, validating that the hip velocity sequence is sufficient to predict contact patterns for diverse motions. Notably, scaling the hip velocity sequence by a single factor before feeding it into the $f_\delta(\cdot)$ still preserves the high prediction precision, indicating that the contact pattern is more closely associated with {the change of hip speed trend} than with speed magnitude, which aligns well with the observation from Fig.~\ref{fig:observations}. The details are shown in the Appendix. 

To control the contact by the hip, our method aims to generate motions adhering to the learned hip-contact relationship by $f_\delta(\cdot)$ to control the contact, synthesizing high-quality results without foot-skating artifacts and can control contact flexibility.

\subsection{Architecture}
\label{sec:transferNet}
\label{sec:architecture}

\begin{figure}[h]
	\centering
	\includegraphics[width=1.0\linewidth]{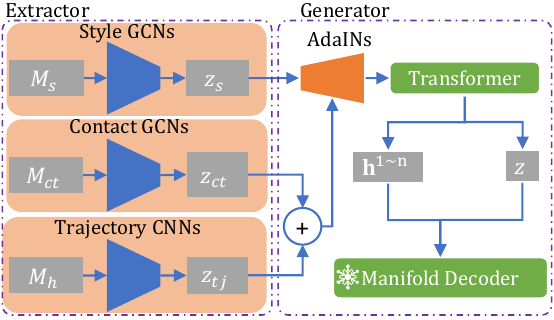}
    \caption{Overview of our pipeline. 
The grey blocks represent the data and others indicate the network.
Trapezium shapes indicate the presence of downsampling and upsampling in the network. The snowflake symbol denotes the manifold decoder is frozen. Note that only the hip velocity of $M_h$ is used as an input to Trajectory CNNs (see Appendix for details).
}
	\label{fig:pipeline} 
 \vspace{-1em}
\end{figure}

\begin{figure}[hb]
	\centering
	\includegraphics[width=1.0\linewidth]{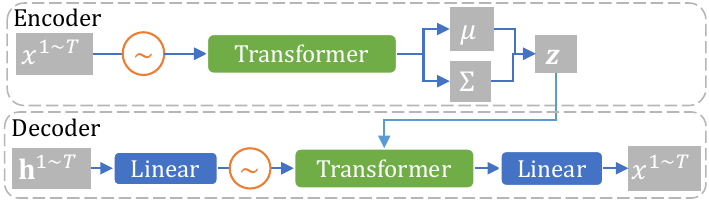}
    \caption{Overview of our CVAE. The \textasciitilde~ in a circle represents the global positional embedding.}
	\label{fig:CVAE} 
 \vspace{-2em}
\end{figure}

As shown in Fig.~\ref{fig:pipeline}, our architecture consists of two stages. In the first stage, we apply three different networks to extract the features of style $z_s$, contact timing  $z_{ct}$ and trajectory $z_{tj}$ from three input motions ($M_s$, $M_{ct}$ and $M_h$), respectively. In the second stage, these three variables are composed to generate motions. {Specifically, the contact timing and trajectory features are composed into a content feature, which is modulated with the style feature by AdaIN blocks. Then a transformer predicts the hip velocity that satisfies the contact and trajectory conditions and a latent variable $z$ for encoding the remaining style variations. Lastly, a manifold synthesizes the intended motion. We employ a conditional variational auto-encoder (CVAE) as the manifold, using the hip as the condition, illustrated in Fig. \ref{fig:CVAE}. Since the relationship is a kind of temporal pattern, our CVAE encodes the randomness of the whole sequence instead of encoding frame-level characteristics as in ~\cite{tang2022cvae,ling_character_2020}. Besides, we apply an attention mechanism that employs hip velocity as the query to emphasize the relationship between hip velocity and motion dynamics. See Appendix for details.}

Our manifold has two advantages. First, the manifold synthesizes motion sequences with leg movements that are compatible with hip velocity, reducing foot skating artifacts. Second, it decouples the hip velocity from the motion, giving us great control over the trajectory and contact timing, as shown in Sec.~\ref{sec:trajectory_editing}.

\subsection{Losses}
Instead of sampling three sequences from the dataset for every training step, we sample a style sequence $M_s$ and a content sequence $M_c$, similar to previous motion style transfer methods. During training, we extract $z_s$ from $M_s$, $z_{tj}$ from $M_c$, and extract $z_{ct}$ alternatively from  $M_s$ and $M_c$.
We utilize the reconstruction loss $L_{\text{rec}}$ and the cycle consistency loss $L_{\text{cyc}}$ as described in \cite{aberman2020unpaired,Jiang_motion_puzzle}:
\begin{align}
L_{\text{cyc}} &= ||G(M_{c},G_{scc},G_{scc}) - M_c||_1   + ||G(G_{scc},M_s,M_s) - M_s||_1,\nonumber \\
L_{\text{rec}} &= ||G_{ccc} - M_c||_1 + ||G_{sss}-M_s||_1, 
\end{align}
where $||\cdot||_1$ represents L1 norm and $G_{sch}=G(M_s,M_c,M_h)$ denotes the synthesized motion with style extracted from $M_s$, contact timing from $M_c$, and trajectory from $M_h$. For instance, $G_{scc}$ represents a motion with the style derived from $M_s$, while maintaining the contact timing and trajectory of $M_c$.
Besides, to separate the trajectory, we integrate a trajectory loss, as indicated by:
\begin{align}
 L_{\text{tj}} &= ||\mathbf{H}_{scc} - \mathbf{h}_{c}||_2^2 + \alpha_{tj} ||proj(\mathbf{H}_{ssc})-proj(\mathbf{h}_{c})||_2^2,
\end{align}
where $\mathbf{H}_{scc}$ denotes the hip velocity of $G_{scc}$, and $proj(\cdot)$ extracts the trajectory by projecting the hip position onto the ground. 
Since both the trajectory and contact timing of $\mathbf{H}_{scc}$ should converge to $\mathbf{h}_c$, the first term does not extract the trajectory explicitly.
In our experiment, we empirically set $\alpha_{tj}=0.2$ to allow for a small deviation in trajectory.
In addition, we propose a contact loss $L_{\text{ctt}}$ to separate the contact timing, as denoted by:
\begin{equation}
\label{eq:loss_ctt}
    L_{\text{ctt}} = ||f_\delta(\mathbf{H}_{scc})-f_\delta(\mathbf{h}_c)||_2^2 +||f_\delta(\mathbf{H}_{ssc})-f_\delta(\mathbf{h}_s)||_2^2, 
\end{equation}
where $f_\delta(\cdot)$ is the learned function that captures the relationship between the hip velocity and the contact timing. 
We further propose a style loss $L_{\text{style}}$ to enhance the style, leveraging the encoder of CVAE:
\begin{equation}
\label{eq:loss_sty}
    L_{\text{style}} = ||g(E(G_{scc}))-g(E(M_{s}))||_2^2, 
\end{equation}
where $E(\cdot)$ represents the latent variable from the final layer of the CVAE encoder, and $g$ denotes the Gram matrix.
As a result, the loss of our architecture is defined as:
\begin{equation}
L =  L_{\text{rec}}+ \alpha_{0} L_{\text{cyc}}+ \alpha_{1} L_{\text{style}}+\alpha_{2} L_{\text{tj}}+\alpha_{3}  L_{\text{ctt}},
\end{equation}
where all $\alpha_{i}$ are empirically set to 0.5 in our experiments.

To train the motion manifold, we employ a $\beta$-VAE training procedure, which aims to minimize both the reconstruction and KL-divergence losses.
Note that our manifold learns the hip-contact relationship by representation learning without explicitly incorporating any related loss.
The implementation details and data formatting are shown in the Appendix.

\section{Style transfer controlling}


\subsubsection*{Trajectory Controlling}
\label{sec:trajectory_editing}
Transferring style without adjusting velocity may introduce unnaturalness if the character moves at a completely different speed in the style sequence than it does in the content sequence.
By scaling the average magnitude of the hip velocity before feeding it into the transformer decoder, our method can modify speed magnitude without affecting contact or style. 
The experiments are shown in Sec.\ref{sec:exp_manifold_cap}.
The results are shown in the Fig.~\ref{fig:scale}.
Besides, animators can conveniently set the trajectory from another motion without impacting the style by replacing the hip sequence input to the manifold decoder with the target hip sequence. This operation also changes the contact timing because it conforms to the hip velocity. The result is depicted in the final image of Fig.~\ref{fig:scale}.
To maintain contact timing, our method allows for the interpolation of trajectory features $z_{tj}$ to establish a gradual transition from one trajectory to another. The results are shown in the third row of Fig.~\ref{fig:teaser} and Fig.~\ref{fig:trajectory_vis}.

\vspace{-0.5em}\subsubsection*{Contact Timing Controlling}
\label{sec:content_editing}
We can automatically edit the contact timing by linearly interpolating  $z_{ct}$ between the content motion and the style motion. As demonstrated in the second row of Fig.~\ref{fig:teaser}, when interpolating from 0 to 1,
the contact timing of the resulting motion 
approaches that of the faster-paced target sequence, resulting in a lighter and more agile style, while the spatial-temporal variations of ``high-knees'' and trajectory remain unchanged.

Our manifold also provides additional flexibility for controlling the contact timing, eliminating the need for reference motions, by solving the following optimization:
\begin{equation}
\label{eq:optimization}
\mathop{\arg\min}_{\mathbf{\hat{h}}}\lambda||proj(\mathbf{\hat{h}})-proj(\mathbf{h})||_2^2 + ||f_\delta(\mathbf{\hat{h}})-c_t||_2^2 + \sigma ||\mathbf{\hat{h}}-\mathbf{h}||_2^2,
\end{equation}
where $c_t\in \mathcal{R}^{T\times 2}$ stands for the desired contact timing, $\lambda$ is the weight for maintaining the trajectory, and $\sigma$ is the weight for the regularization term. We begin the optimization by setting the initial value to the original hip velocity. The optimization aims to find the hip velocity that maintains approximately the existing trajectory while achieving the desired contact timing. After finding the intended hip velocity, our manifold modifies the leg movements accordingly. In Fig.~\ref{fig:contact_edit}, we demonstrate the use of this method to add footsteps and switch legs.
\vspace{-0.5em}\subsubsection*{Style Controlling}
Like previous methods~\cite{aberman2020unpaired, Jiang_motion_puzzle}, interpolating between two style features $z_s$ can affect the spatial-temporal variations of the resulting motion, as shown in the first row of Fig.~\ref{fig:teaser}.

\section{Experiments and Results}
\label{sec:experiments}
This section proposes multiple metrics (Sec.~\ref{sec:manifold_metric}) and demonstrates ablation study (Sec.~\ref{sec:exp_manifold}) and comparisons (Sec.~\ref{sec:exp_style_transfer}). {All evaluations are based on human animations represented by joint rotations, rather than network output. Human animation reconstruction can be divided into three categories based on the output data used: rotation-based, position-based, and velocity-based, which primarily use rotations, positions, and velocities. The Appendix shows details of these reconstruction methods and a related ablation study. 
} We choose the velocity-based way for our method because it produces smooth motion and reduces foot-skating artifacts. 
{However, because previous methods may not produce reasonable velocity and the velocity-based method will degrade their performance, we choose the most appropriate method for each to allow for fair comparisons.}

\vspace{-0.5em}
\subsection{Metrics}

\label{sec:manifold_metric}



\subsubsection{Manifold Metrics}
{We compare our manifold to three recent manifolds: MLD ~\cite{chen2023executing}, VQVAE~\cite{zhang2023generating}, and MVAE~\cite{ling_character_2020}. In addition, we propose a variant of our model that encodes frame-level randomness rather than sequence-level randomness, with architecture details provided in the Appendix. We set MLD's hyperparameters to be identical to our method, with the exception of the hip condition.  For MVAE, we used a variant~\cite{tang2022cvae} that treats the hip as a condition. We also included the hip as a condition in VQVAE~\cite{zhang2023generating} to ensure a fair comparison.
}

We measure the manifolds from two aspects: contact timing controllability and motion quality. 
To this end, we randomly sample $20$ motion sequences from each manifold for each hip trajectory. {We employ the precision and recall rate of the predicted contact to evaluate controllability, which we refer to as \textit{Contact Precision-Recall}. Precision is the percentage of the predicted contact change frames (feet touch or lift off the ground) that correctly match the ground-truth contact changes, while recall is the percentage of ground-truth contact changes that correctly match predicted contact changes.}
When there is no ground truth contact change, we apply $f_\delta({\mathbf{h}})$ for replacement, where $\mathbf{h}$ is the predicted hip velocity. These metrics are distinguished by an asterisk (*) in the upper right corner of the results in the tables.
To evaluate motion quality, we use three metrics. The first is the foot skating metric~\cite{zhang_mode-adaptive_2018}, which calculates the average foot velocity $v_f$ when the foot height $h$ is within a threshold ($H=2.5$), as defined by $L_f = v_f\cdot \text{clamp}(2-2^{{h}/{H}},0,1)$.
The second metric is FMD, which calculates the distance between the distribution of the dataset and the distribution of the randomly sampled data. We employ the joints' velocity to calculate the mean and covariance of the distribution. 
Finally, we conduct a user study in which $20$ participants, five of whom are professional animation designers, evaluate the naturalness of motion sequences generated by different manifolds. For each manifold, we randomly sample five sequences with the same hip trajectory, and sample five different hip trajectories in total, resulting in 25 sequences for each manifold.
Participants assign scores ranging from $1$ to $5$ concerning naturalness, with less than $3$ denotes unnaturalness and more than $3$ indicates naturalness. A score of $3$ denotes indistinguishability.
\vspace{-0.5em}
\subsubsection{Style Transfer Metrics}
\label{subsbusec:style_transfer_metrics}
We evaluate the controllability of trajectory, contact timing, and style, respectively. Contact timing controllability is measured using contact precision-recall, while trajectory accuracy is the L2 distance between the trajectories of ground-truth motions and those of the generated motions. The style is measured by the FMD and style recognition accuracy (SA), similar to~\cite{Jiang_motion_puzzle}. 

\subsection{Ablation study}
\label{sec:exp_manifold}

The experimental results on STYLE100 are shown in Tab.~\ref{tab:Manifold} and we leave the results on CMU in the Appendix.

\subsubsection{Manifold quality and controllability}
\label{subsec:comparision_of_manifold_quality}
Among these manifolds, our manifold demonstrates standout performance, particularly concerning human perception scores on the STYLE100 dataset shown in Tab.~\ref{tab:Manifold} (the score of $4$ represents the naturalness and we achieve $4.46$). Besides, our method consistently achieves high ranks in terms of FMD and foot skating metrics. 
\label{subsec:comparison_of_controllability}
In terms of contact precision and contact recall rate, indicated by Ct P. and Ct R. in 
Tab.~\ref{tab:Manifold}, our method outperforms two other transformer-based networks, MLD and Frame-Level. MLD does not explicitly condition hip velocity while Frame-Level applies the frame-level encoding rather than sequence-level encoding. These experiments validate the significance of the hip condition and long-clip encoding for learning the long-term relationship between contact timing and hip velocity.
In addition, the effect of long-clip encoding for learning the relationship is validated by various architectures. VQVAE, which primarily employs convolution layers, achieves comparable contact precision-recall to our method as it also encodes the long sequence characteristic by down-sampling. In contrast, MVAE, which applies a Mixture-of-Experts MLP and uses frame-level encoding, exhibits lower contact precision-recall.



\subsubsection{Contact precision-recall for motion quality evaluation}
\label{subsec:Contact_precision_metric_for_motion_quality}
The contact precision-recall is more meaningful in evaluating motion quality than the common metrics: the foot skating and the FMD. 
{In particular, the foot skating metric only takes into account the foot velocity and can achieve low error by anticipating unintended zero value.}
Besides, the FMD metric evaluates distribution similarity and may not adequately represent quality, either.
{In contrast, the contact precision-recall indirectly evaluates the extent to which a sequence satisfies the hip-contact relationship of human motion, reflecting certain types of motion naturalness properly.}
As demonstrated in Tab.~\ref{tab:Manifold}, high contact precision-recall metrics consistently correspond to high perception scores while FMD and foot skating do not.
Specifically, VQVAE performs poorly in terms of FMD but ranks among the top-performing methods in terms of human perception, which indicates the inaccuracy of the FMD metric in measuring motion quality. Furthermore, as most of the manifolds achieve the foot skating metric that is close to the metric of the dataset ($0.25$), the foot skating metric is not sufficiently discriminative in this case. 

\subsubsection{Contact controllability and style}
\label{sec:manifolds_style}

\begin{table}[ht]
\tiny
\centering
\caption{Comparisons between our manifold with three previous manifolds and a variation of our manifold. 
"Percep." represents the human perception score. Fv of the dataset is $0.68$.}
\label{tab:Manifold}
\resizebox{1.0\columnwidth}{!}{%
\begin{tabular}{lllllll}

\specialrule{0em}{1pt}{0pt} \multicolumn{1}{l}{Methods} &
  \multicolumn{1}{l}{Ct P.} &
  \multicolumn{1}{l}{Ct R.} &
  \multicolumn{1}{l}{${\text{FMD}_{\text{vel}}}$} &
  \multicolumn{1}{l}{Fv} &
  \multicolumn{1}{l}{Percep.} &
  \\ \hline
STYLE100 &&&&\\\hline
MLD  &0.4618* & 0.4311*  &0.0265 &0.92 &N/A \\
MVAE &0.6129 &0.4843  &\textbf{0.0116}&0.84&1.83\\
{Frame-Level}  &0.5613 &0.6480  &0.0190 & 0.95 &2.40\\
VQVAE &\underline{0.8633} &\underline{0.8554}  &0.0475 & \textbf{0.67} &\underline{3.83}\\
Ours &\textbf{0.8782} &\textbf{0.8712}  & \underline{0.0157} &\underline{0.79} & \textbf{4.46} 
\end{tabular}}
\vspace{-1em}
\end{table}

This section evaluates the performance when replacing manifolds in our style transfer framework with other manifolds. We conduct three experiments to evaluate the style effects and controllability of contact timing and trajectory, respectively. We use Motion Puzzle~\cite{Jiang_motion_puzzle} as a baseline for style effects. The results are shown in Tab.~\ref{tab:manifold_style}. 

The prerequisite for applying $L_{\text{style}}$ (Eq.~\ref{eq:loss_sty}) is that the manifold must decouple the style variations from the hip velocity, so that constraining latent variable $z$ does not affect contact timing and trajectory. Otherwise, adding $L_{\text{style}}$ significantly degrades performance, as demonstrated in the Appendix.
Therefore, we do not employ $L_{\text{style}}$ to ensure fair comparisons.

Our manifold achieves the best scores for expressing style, as seen in the FMD and SA metrics for all three experiments in Tab.~\ref{tab:manifold_style}. In the style experiment, other manifold methods tend to prioritize the reconstruction of the content sequence rather than conveying style, resulting in high contact precision-recall but low FMD and SA. We attribute our superior performance to the separation of trajectory and contact timing from spatial-temporal variations of style, so constraining contact timing and trajectory did not significantly affect the style.

Capturing the hip-contact relationship is critical for contact controllability. As evidence, MLD, MVAE, and Frame-Level, which have not learned the relationship, exhibit lower contact precision-recall than our manifold in contact controllability experiments. 
Notably, the methods that preserve the hip-contact relationship, such as VQVAE and our manifold, encounter the challenge of modifying the trajectory or contact timing without alerting each other because both factors are relevant to hip velocity.
As demonstrated in the trajectory and contact experiment in Tab.~\ref{tab:manifold_style}, VQVAE struggles in resolving conflicts between meeting the requirements of contact timing and trajectory, which causes both factors to deviate from the intended target, leading to the worst results compared to other methods.
Nevertheless, our proposed manifold satisfies contact timing and trajectory requirements effectively, achieving the best contact precision-recall among all the manifolds. In addition, the resulting trajectory differences (8.9 cm) for a two-second sequence are hard to discern visually.

\begin{table}[ht]
\tiny
\centering
\caption{Comparisons between different manifolds for our framework. (XZ, Angle) are trajectory metrics, (FMD, SA) are style metrics and (Ct P., Ct R.) are contact timing metrics. Fv represents the foot skating metric. 
Fv of the dataset is $0.64$.}
\label{tab:manifold_style}
\resizebox{1.0\columnwidth}{!}{%
\begin{tabular}{llllllllll}
\specialrule{0em}{1pt}{0pt}
\multicolumn{1}{l}{Methods} &
\multicolumn{1}{l}{(XZ}  &
\multicolumn{1}{l}{Angle)} &
\multicolumn{1}{l}{(FMD} &
\multicolumn{1}{l}{SA)}&
\multicolumn{1}{l}{(Ct P.} &
\multicolumn{1}{l}{Ct R.)} &
\multicolumn{1}{l}{Fv}
\\ \hline
Style &  &  &&&&&  \\  \hline
\specialrule{0em}{1pt}{0pt}
Motion Puzzle & 5.5  & 0.046 & 87 & \textbf{0.920} &0.467&0.476  & 1.68 \\
MLD & 3.3 & 0.027& 135  & 0.751 & 0.862&0.872 & \textbf{0.60}  \\
MVAE & 3.1 & 0.017& 157  & 0.763 &0.840&0.817& 0.99\\
Frame-Level & 2.5 & 0.027 &194 & 0.571&\textbf{0.919}&\textbf{0.926}& 0.61 \\
VQVAE & 4.1 & 0.032 & 182  & 0.603&0.861&0.876&0.53 \\
Ours & \textbf{2.4} & \textbf{0.013} & \textbf{85}  &{0.879} &0.849 &0.849  &0.61 \\
\hline
Contact\\
\hline
MLD & \textbf{5.2} &0.039 &105 &0.847 &0.586 &0.593&0.82\\
MVAE & 8.89 & 0.030& 128  & 0.824 &0.649 &0.645& 0.97\\
Frame-Level & 8.4 &0.037 &100 &0.847 &0.610 &0.651& 0.64 \\
VQVAE & 13.4 & 0.046 & 115 & 0.806&0.575 &0.586& 0.64  \\
Ours &8.9 &\textbf{0.031} &\textbf{70} &\textbf{0.931} &\textbf{0.741}&\textbf{0.784}&\textbf{0.58}   \\
\hline
Trajectory\\
\hline
MLD & \textbf{6.1} &0.037 &129 &0.805 &0.561 &0.566&0.62 \\
MVAE & 8.1 & \textbf{0.027}& 260 & 0.239&0.593 &0.571& 0.87 \\
Frame-Level & 7.7 &0.029 &159 &0.734 &0.590 &0.651&0.63\\
VQVAE & 13.9 & 0.047 & 143 & 0.719&0.531 &0.569& \textbf{0.50}  \\
Ours &8.9 &0.031 &\textbf{65} &\textbf{0.943} &\textbf{0.642} &\textbf{0.678}&0.60\\
\end{tabular}}
\vspace{-3em}
\end{table}

\subsection{Comparisons}
\label{sec:exp_style_transfer}

We evaluate our method against previous style transfer methods, including \cite{aberman2020unpaired}, Motion Puzzle~\cite{Jiang_motion_puzzle}, and a variant of Motion Puzzle that utilizes the similar decoupling formulation (denoted as Motion Puzzle ($+$ decouple)). 
Instead of using the velocity-based way as our method, ~\cite{aberman2020unpaired} and Motion Puzzle employ the rotation-based way to generate the final animation because \cite{aberman2020unpaired} uses rotation representation only and {it's difficult for Motion Puzzle to accurately learn the joints' velocities.}
Additionally, \cite{aberman2020unpaired} sets hip velocity using a heuristic solution and does not preserve the trajectory. Therefore, we omit trajectory metrics (xz and angle) for \cite{aberman2020unpaired}. 

We conduct three experiments for evaluating the controllability of style, trajectory and contact timing by setting the corresponding interpolation factor to 0.5 and 1.0, respectively. The style controllability experiment with an interpolation factor of 1 is equivalent to the previous motion style transfer, which can validate that our method not only outperforms previous methods in controlling contact but also maintains superior style effects. We omit the results of the interpolation of 0.5 in Tab.~\ref{tab:style} because it results in a similar conclusion as the interpolation of 1.0. The complete table is shown in the Appendix.

\begin{table}[ht]
\tiny
\centering
\caption{Comparisons between our method with previous motion style transfer methods. 
}
\label{tab:style}
\resizebox{1.0\columnwidth}{!}{%
\begin{tabular}{llllllllll}
\specialrule{0em}{1pt}{0pt}
\multicolumn{1}{l}{Methods} &
\multicolumn{1}{l}{(XZ}  &
\multicolumn{1}{l}{Angle)} &
\multicolumn{1}{l}{(${\text{FMD} }$} &
\multicolumn{1}{l}{${\text{SA}}$)}&
\multicolumn{1}{l}{(Ct P.} &
\multicolumn{1}{l}{Ct R. )} &
\multicolumn{1}{l}{Fv} \\
\hline
Style \\
\hline
Aberman et al. &N/A &N/A &191 &0.732 &\textbf{0.791}  &\underline{0.773}  &\underline{0.83} \\
Motion Puzzle & 5.5 & 0.046 & 87  & 0.920&0.467 &0.476 &1.68 \\
$+$ decouple& \underline{1.7} & \textbf{0.010} & \textbf{69}   & \underline{0.940} &0.363  &0.294  &1.91  \\
Ours &\textbf{1.6} &\underline{0.014} &\underline{72} &\textbf{0.943} &\underline{0.782} &\textbf{0.799} &\textbf{0.63} \\
\hline
Contact \\
\hline
Aberman et al. &N/A &N/A &180 &0.773 &0.232 &0.498 &2.05 \\
Motion Puzzle & 79 & 0.862 & \textbf{40}  & \textbf{0.990} &\textbf{0.874} &\textbf{0.919} &\underline{0.72}\\
$+$ decouple & \underline{3.6} & \textbf{0.020} & \underline{52} & \underline{0.968} &0.458 &0.270 &1.91\\
Ours &\textbf{2.9} &\underline{0.031} &70 &0.931 &\underline{0.741} &\underline{0.784} &\textbf{0.58} \\
\hline
Trajectory \\
\hline
$+$ decouple & \textbf{2.7} & \textbf{0.015} & \textbf{48} & \textbf{0.970} &0.400 &0.510 &1.46\\
Ours &8.9 &0.031 &65 &0.943 &\textbf{0.642} &\textbf{0.678} &\textbf{0.60}\\
\end{tabular}}
\vspace{-3.em}
\end{table}

\begin{table}[ht]
\tiny
\centering
\caption{The performance of methods using IK post-processing.}
\label{tab:style_post_processing}
\resizebox{1.0\columnwidth}{!}{%
\begin{tabular}{llllllll}
\specialrule{0em}{1pt}{0pt}
\multicolumn{1}{l}{Methods} &
\multicolumn{1}{l}{(${\text{FMD} }$} &
\multicolumn{1}{l}{${\text{SA}} $)}&
\multicolumn{1}{l}{(Ct P.} &
\multicolumn{1}{l}{Ct R. )} &
\multicolumn{1}{l}{Fv } \\
\hline
Style \\
\hline
Aberman et al.  & 247 & 0.574 &0.839 & 0.798 & 0.67 \\
Motion Puzzle  &  167 &  0.760 &0.800 & 0.866 & 0.78 \\
$+$ decouple &  145  & 0.812 & 0.733 &0.867 & 0.87 \\
\end{tabular}}
\vspace{-2em}
\end{table}

\subsubsection{Style effects and motion naturalness}
Our method achieves almost the best results for conveying style (FMD, SA) while also preserving the motion's naturalness.
%
Specifically, in terms of motion naturalness, 
Motion Puzzle fails to generate high-quality results, as indicated by the high foot skating value and low contact precision-recall. Although~\cite{aberman2020unpaired} improves the quality by employing discriminators, it still has worse foot skating artifacts than our method.
To reduce foot skating, both ~\cite{aberman2020unpaired} and Motion Puzzle employ contact-based IK solvers as post-processing to preserve the contact timing of the content sequence. However, applying this post-processing may undermine the style, as indicated by the increased FMD and decreased SA values presented in Tab.~\ref{tab:style_post_processing}. In section~\ref{sec:exp_manifold_cap}, we show that projecting their results onto our manifold is a better way for post-processing.


\subsubsection{Contact controllability}
Motion Puzzle's interpolation with a factor of 1.0 is equivalent to reproducing the style sequence, thus easily achieving superior contact precision-recall and style metrics (FMD and SA). However, it devastates the trajectory requirements, demonstrated by the trajectory error of up to 79 cm for factor 1.0. 
Decoupling the trajectory enables achieving trajectory requirements, but it can lead to leg movement incompatible with velocity, resulting in poor contact precision-recall metrics.
Overall, our method modifies the contact timing while barely affecting style or trajectory, exhibiting the best contact controllability among the compared methods.

\subsubsection{Trajectory controllability}
Neither~\cite{aberman2020unpaired} nor Motion Puzzle decouples the trajectory, so we did not compare them on the trajectory controllability task. Although the ($+$ decouple) approach achieves a trajectory closer to the desired trajectory, this is accomplished by ignoring the contact timing completely, evidenced by the poor performance in Ct P. and Ct R. Synthesizing a motion with a significantly different trajectory but similar contact timing is a challenging task. Our method performs better in this regard. 
\subsection{Manifold Capability}
\label{sec:exp_manifold_cap}
The manifold capability is evaluated from three perspectives.
First, our manifold is able to preserve contact timing even when scaling the magnitude of the hip speed.
Second, our approach allows for the complete replacement of both the trajectory and contact timing without compromising the quality of motion or style.
Lastly, our proposed manifold approach can also serve as a post-processing stage for previous style transfer methods, improving the motion quality and providing additional control capability.

All the experimental results are presented in Tab.~\ref{tab:manifold_cap}, with the original style transfer results ("Original" in the table) serving as a baseline. Our manifold demonstrates high contact precision-recall in scaling experiments, proving its ability to scale speed magnitude while preserving contact timing. Additionally, after replacing the hip velocity, our manifold achieves the desired trajectory and contact timing while maintaining style.

In addition, both scaling and replacing operations are relevant to the manifold only. Therefore, even without applying our complete approach, the contact timing and trajectory can be controlled by projecting a motion onto our manifold. 
To evaluate the performance of the projection, we project the results from previous style transfer methods. This projection approach significantly improves the motion quality, with fewer foot skating artifacts when compared to the results in Tab.~\ref{tab:style} and better style effects than when using a contact-IK based solver as the post-processing. In conclusion, our manifold can serve as post-processing for other style transfer methods, improving motion quality while providing contact controllability.

\begin{table}[ht]
    \tiny
    \centering
    \caption{Manifold capability experiments comprise editing the hip velocity and projecting previous transfer methods' results onto our manifold.}
    \label{tab:manifold_cap}
    \resizebox{1.0\columnwidth}{!}{%
        \begin{tabular}{lllllllll}
            \specialrule{0em}{1pt}{0pt} 
            \multicolumn{1}{l}{Operation} &
            \multicolumn{1}{l}{(XZ }  &
            \multicolumn{1}{l}{Angle )} &
            \multicolumn{1}{l}{(${\text{FMD} }$} &
            \multicolumn{1}{l}{${\text{SA}} $)}&
            \multicolumn{1}{l}{(Ct P. } &
            \multicolumn{1}{l}{Ct R. )} &
            \multicolumn{1}{l}{Fv}
            \\ \hline
            Hip editing \\
            \hline
            Original  &2.6 &0.0190  &66 &0.940 &0.773 &0.787&0.61\\
            Scaling & 0.62 & 0.0065   & 59  & 0.941 &0.760 &0.763& 0.86\\
            Replacing & 0.44 & 0.0044 & 20  & 0.992 &0.927 &0.908 & 0.57 \\
            \hline
            Projection\\
            \hline
            Aberman et al. &4.2&0.074 &189 & 0.670 &0.882 &0.830 &0.41\\
            Motion Puzzle &5.4 & 0.046   & 105  & 0.842 & 0.832 &0.882 & 0.50\\
            $+$ decouple &1.2 & 0.009  & 84  & 0.880 &0.794 &0.839  & 0.53\\
        \end{tabular}}
    \vspace{-2.em}
\end{table}
\section{Discussions}
\subsection*{Phase-related methods}
{
Previous methods \cite{starke_neural_2021,starke_local_2020,holden2017phase} can use phase variables to control the contact. They first extract contact-related intermediate variables from motions and then learn the motion distribution based on the intermediate variables.  While phase-related methods provide precise control over contact, they have not fully explored the variety of styles. For example, if the phase condition constrains multiple body-part movements, the result may be limited to a specific style  \cite{tang2023rsmt}. Furthermore, these methods model phase using a temporal continuous function and frequently require auto-regressive manner prediction from previous frames, reducing style diversity even further as the style is influenced by previous frames. 

Unlike previous phase-related methods, we use hip velocity to control contact.  Furthermore, we model the intrinsic relationship of human motion without imposing any explicit artificial constraints. In the absence of an explicit constraint, we initially struggled to generate precise out-of-distribution contact timing because we couldn't find a suitable hip velocity as the condition. However, we believe it is appropriate for a style transfer task because rhythm and frequency, which are easier to achieve, can express the style to some extent. Explicitly adhering to out-of-distribution contact patterns \cite{starke_local_2020} may reduce motion quality and is undesirable for motion style transfer.
}
\subsection*{Limitations and Future Work}


If the desired contact and trajectory are incompatible, our method may produce an unnatural motion. For example, transferring a walking motion to one that does not rely on the foot to support and move may result in a floating motion (see the failure case in Fig.~\ref{fig:failure_case} and our video).
 However, as shown in our video, the relationship holds for diverse motions including locomotion, dancing, jumping, and so on, which are applicable to the vast majority of human daily motion types. In addition, even for the cases in which the contact control is not applicable, compared to previous methods, our method is still able to provide extra trajectory control and a similar style transfer because of our manifold's diversity. Future research could explore a more general control method.



Additionally, manifold-based motion generation approaches have become popular in recent days. Lots of studies have utilized diffusion networks with either VAE~\cite{chen2023executing} or VQVAE~\cite{Ao2023GestureDiffuCLIP,zhang2023generating,jiang2023motiongpt} as the manifold. Future research could involve incorporating our manifold and metric to enhance motion quality and editing flexibility in diffusion applications.


\section{Conclusion}

We have presented a novel approach to character motion style transfer that addresses the difficult task of decoupling contact from motions. 
By extracting style, contact timing, and trajectory features, our approach enables fine-grained control over these aspects independently, resulting in more expressive and natural motion. Modeling the relationship between hip velocity and contact timing using a transformer architecture is the key insight for achieving finer control over contact timing while preserving naturalness. We also propose a new metric for measuring the match between the synthesized contact and hip velocity, which is also closely aligned with human perception in terms of motion naturalness. Furthermore, our proposed manifold is versatile in the sense that it can both directly generate motions and be used as post-processing for existing motion transfer techniques. Experiment results show that our method produces high-quality and expressive results for a wide range of motion styles, outperforming state-of-the-art methods in style expressivity and motion quality.

\begin{acks}
Xiaogang Jin was supported by Key R\&D Program of Zhejiang (No. 2023C01047) and the National Natural Science Foundation of China (Grant No. 62472373). We extend our appreciation to the participants who generously devoted their time to our user study.

\end{acks}
\bibliographystyle{ACM-Reference-Format}
\bibliography{sample-base}

\appendix


\begin{figure*}
\centering
  \includegraphics[width=\textwidth]{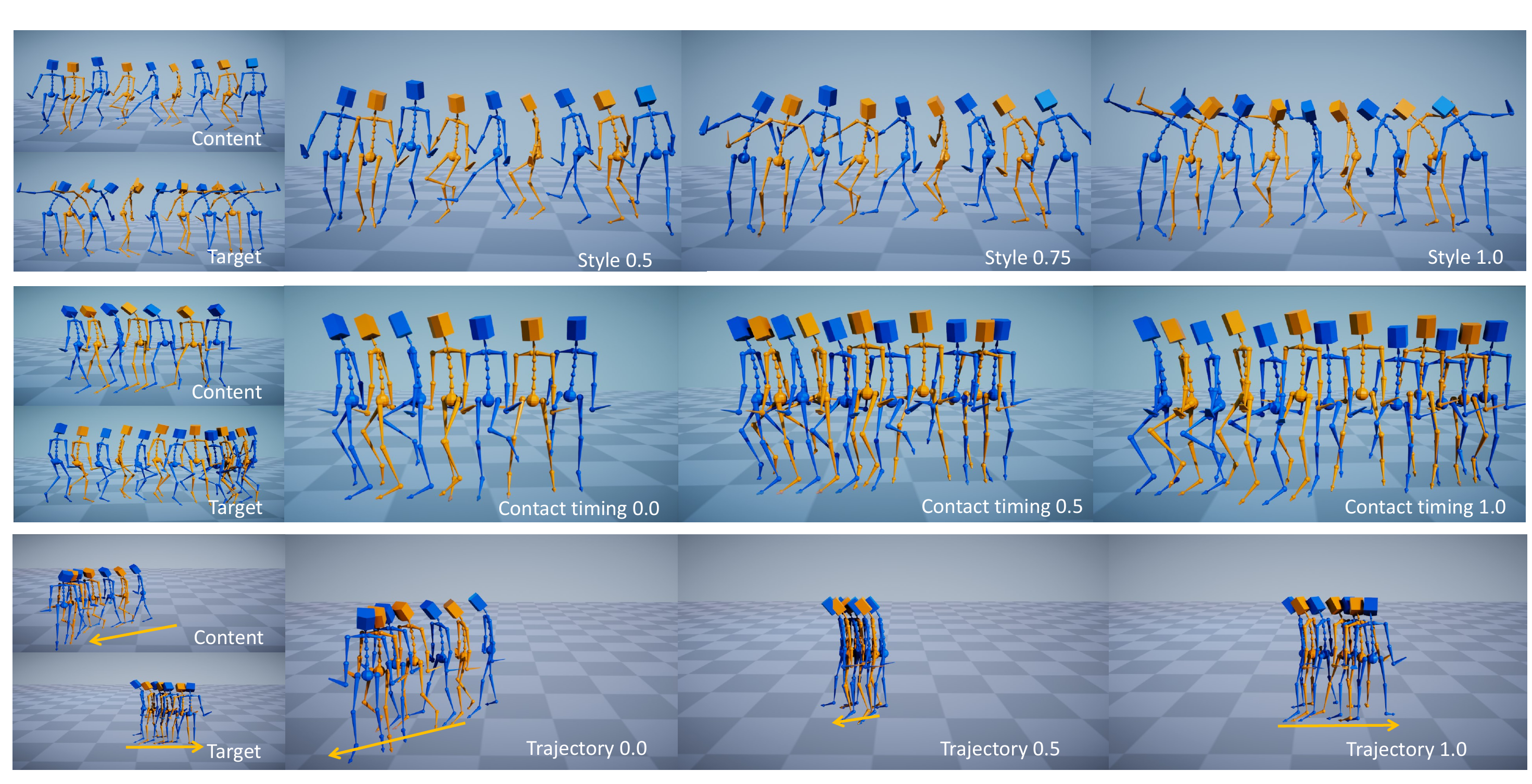}
  \caption{
Our method allows users to control trajectory, contact timing, and style of motions separately or jointly. To improve visibility, we intentionally increase the spatial distance between two adjacent skeletons. Blue skeletons represent the frames when the character makes contact with the ground, while orange skeletons represent the midpoint between two blue frames. Our method allows for separate interpolation of style (first row), contact timing (second row), and trajectory (third row). The latent style/contact/trajectory space interpolation parameter here varies from content sequence (0.0) to target sequence (1.0).
  }
  \Description{}
  \label{fig:teaser}
\end{figure*}

\begin{figure*}
\centering
  \includegraphics[width=\textwidth]{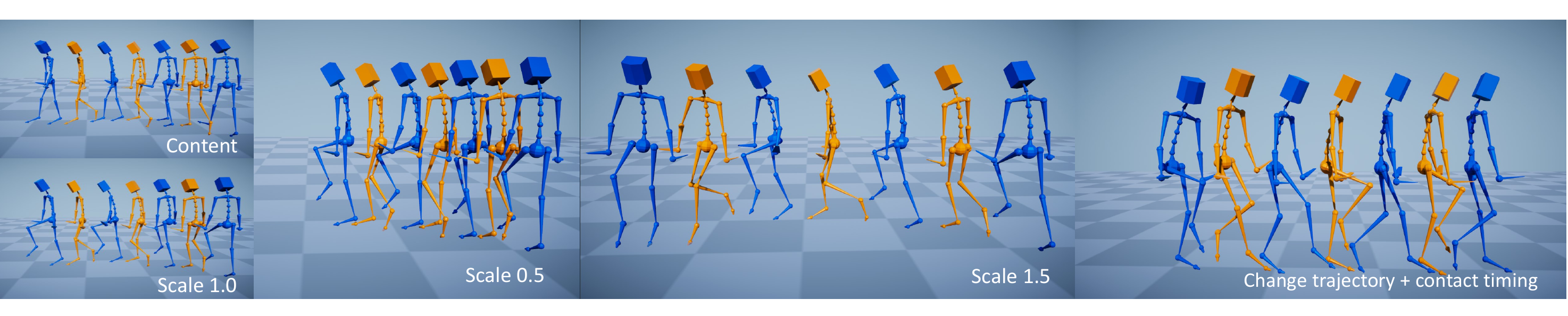}
  \caption{
Our method allows changing the trajectory by scaling the magnitude of the hip velocity (Scale 0.5, Scale 1.0, and Scale 1.5), as well as changing the trajectory and contact timing at the same time by setting its hip velocity from another motion (change trajectory + contact timing).  In this setting, all results are transferred to the "high knees" style.
  }
  \Description{}
  \label{fig:scale}
\end{figure*}

\begin{figure*}
\centering
  \includegraphics[width=\textwidth]{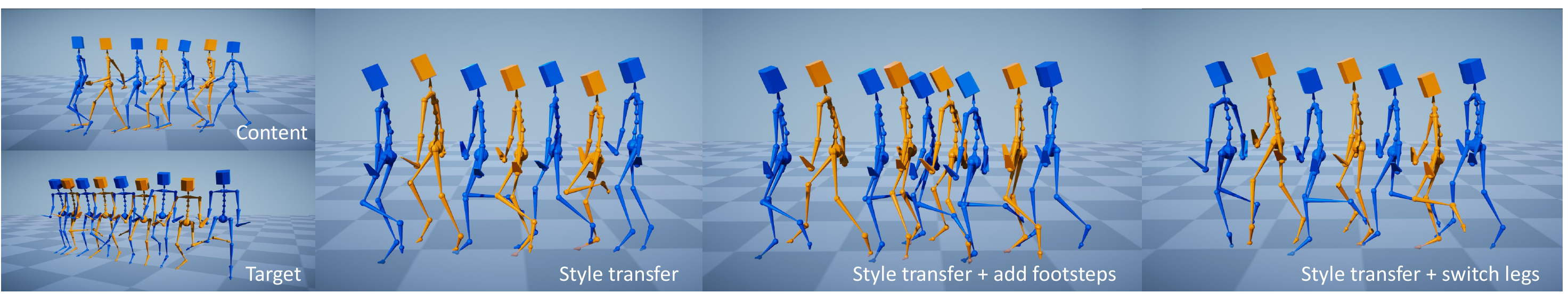}
  \caption{
Our method allows for fine-grained manual control of the contact timing patterns. Following the style transfer process (second column), we show examples of additional control by adding footsteps (second to last column) or switching legs (last column).
  }
  \Description{}
  \label{fig:contact_edit}
\end{figure*}

\begin{figure*}
\centering
  \includegraphics[width=\textwidth]{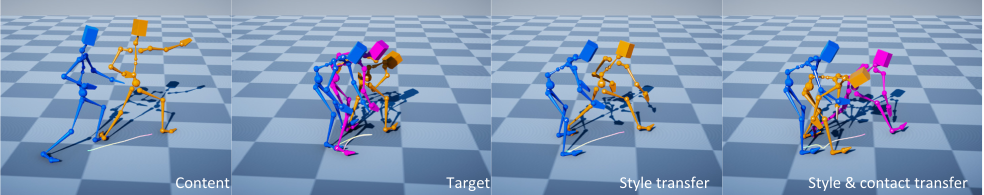}
  \caption{
Our contact control can also be employed in non-locomotion cases. This figure shows a martial art case. The color of the character changes from blue to orange and to purple over time. The character in the content sequence takes a step forward into a bow stance. In the target sequence, the character keeps the left leg static and moves the right leg twice. Our method can transfer the style and contact from the target sequence to the content sequence.
  }
  \Description{}
  \label{fig:martial_arts}
\end{figure*}

\begin{figure*}
\centering
  \includegraphics[width=\textwidth]{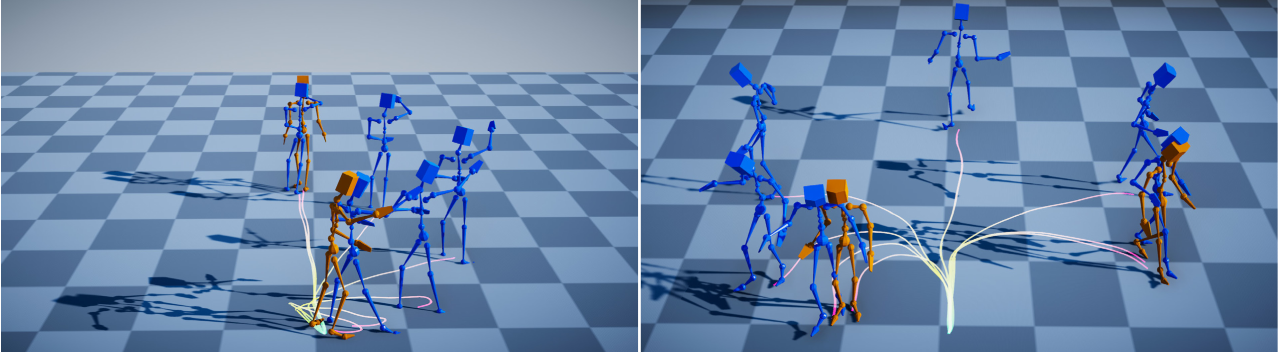}
  \caption{
We show our trajectory interpolation results. The orange characters in each image represent the last frame of the content and target motions, respectively. The blue characters are the last frame of the interpolated motions. The first image demonstrates the interpolation between forward locomotion and a dance motion with intricate trajectory. The second image showcases the interpolation between two motions involving walking in different directions.
  }
  \Description{}
  \label{fig:trajectory_vis}
\end{figure*}

\begin{figure*}
\centering
  \includegraphics[width=\textwidth]{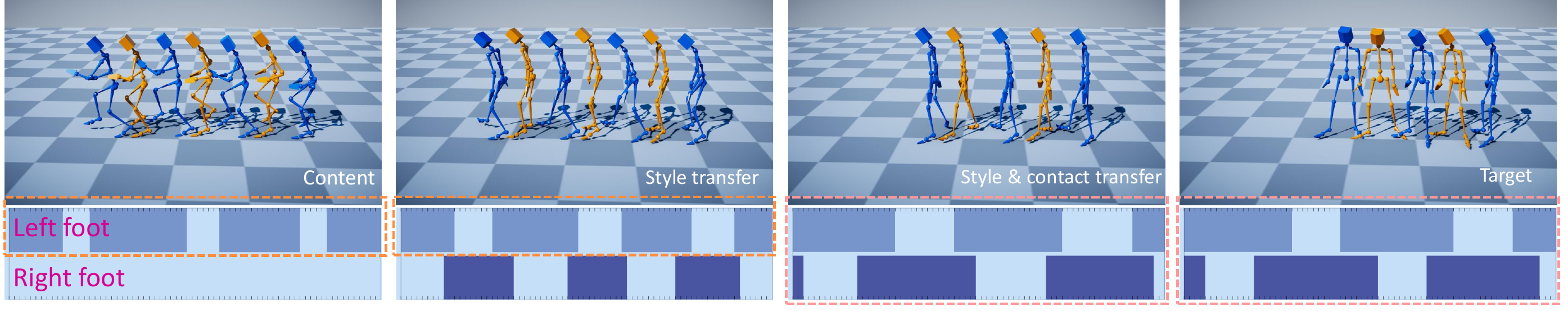}
  \caption{
This case presents our contact control when transferring style from a motion that makes contact using only the left foot to a motion with a "duck foot" style. The style transfer result showcases the "duck foot" style while preserving the contact timing pattern of the left foot, and the style \& contact transfer modifies the contact as well. Similar contact timing patterns are highlighted using the same color boxes. Blue skeletons represent the frames when the left foot of the character makes contact with the ground, while orange skeletons represent the midpoint between two blue frames.
  }
  \Description{}
  \label{fig:lefthop}
\end{figure*}
\begin{figure*}
\centering
  \includegraphics[width=\textwidth]{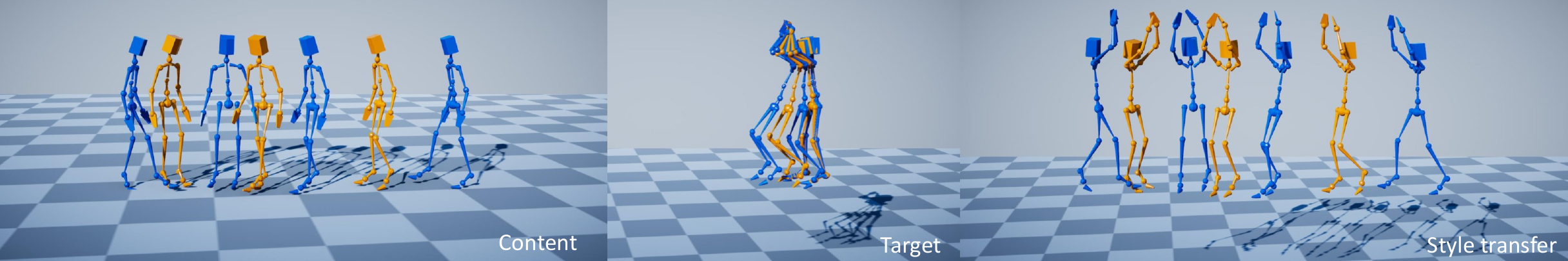}
  \caption{
A failure case. Transferring a walking motion to one that does not rely on the foot for support and movement may result in a floating motion.
  }
  \Description{}
  \label{fig:failure_case}
\end{figure*}
\end{document}